\documentclass[pdflatex,sn-basic,iicol]{sn-jnl}

\jyear{2022}%

\theoremstyle{thmstyleone}%
\theoremstyle{thmstyletwo}%

\theoremstyle{thmstylethree}%

\begin{document}

\title[Human-Centric OOD]{Rethinking Out-of-Distribution Detection From a Human-Centric Perspective}


\author[1]{\fnm{Yao} \sur{Zhu}}\email{ee$\_$zhuy@zju.edu.cn}

\author[2]{\fnm{Yufeng} \sur{Chen}}\email{yuefeng.chenyf@alibaba-inc.com}
\author[2]{\fnm{Xiaodan} \sur{Li}}\email{fiona.lxd@alibaba-inc.com}

\author[2]{\fnm{Rong} \sur{Zhang}}\email{stone.zhangr@alibaba-inc.com}

\author[2]{\fnm{Hui} \sur{Xue}}\email{hui.xueh@alibaba-inc.com}

\author[1,4]{\fnm{Xiang} \sur{Tian}}\email{rongxinj@zju.edu.cn}

\author*[1,4]{\fnm{Rongxin} \sur{Jiang}}\email{tianx@zju.edu.cn}

\author*[3,4]{\fnm{Bolun} \sur{Zheng}}\email{blzheng@hdu.edu.cn}

\author[1,5]{\fnm{Yaowu} \sur{Chen}}\email{cyw@mail.bme.zju.edu.cn}

\affil*[1]{\orgname{Zhejiang University}, \orgaddress{\city{Hangzhou}, \postcode{310027}, \state{Zhejiang}, \country{China}}}

\affil[2]{\orgdiv{The Security Department}, \orgname{Alibaba Group}, \orgaddress{\city{Hangzhou}, \postcode{310023}, \state{Zhejiang}, \country{China}}}

\affil[3]{\orgname{Hangzhou Dianzi University}, \orgaddress{\city{Hangzhou}, \postcode{310018}, \state{Zhejiang}, \country{China}}}

\affil[4]{\orgdiv{Zhejiang Provincial Key Laboratory for Network Multimedia Technologies}, \orgaddress{\city{Hangzhou}, \postcode{310027}, \state{Zhejiang}, \country{China}}}

\affil[5]{\orgdiv{Zhejiang University Embedded System Engineering Research Center}, \orgname{Ministry of Education of China}, \orgaddress{\city{Hangzhou}, \postcode{310027}, \state{Zhejiang}, \country{China}}}


\abstract{\textbf{O}ut-\textbf{O}f-\textbf{D}istribution (OOD) detection has received broad attention over the years, aiming to ensure the reliability and safety of deep neural networks (DNNs) in real-world scenarios by rejecting incorrect predictions. 
However, we notice a discrepancy between the conventional evaluation vs. the essential purpose of OOD detection. On the one hand, the conventional evaluation exclusively considers risks caused by label-space distribution shifts while ignoring the risks from input-space distribution shifts. On the other hand, the conventional evaluation reward detection methods for not rejecting the misclassified image in the validation dataset. However, the misclassified image can also cause risks and should be rejected. 
We appeal to rethink OOD detection from a human-centric perspective, that a proper detection method should reject the case that the deep model's prediction mismatches the human expectations and adopt the case that the deep model's prediction meets the human expectations. 
We propose a human-centric evaluation and conduct extensive experiments on 45 classifiers and 8 test datasets. We find that the simple baseline OOD detection method can achieve comparable and even better performance than the recently proposed methods, which means that the development in OOD detection in the past years may be overestimated. Additionally, our experiments demonstrate that model selection is non-trivial for OOD detection and should be considered as an integral of the proposed method, which differs from the claim in existing works that proposed methods are universal across different models.}

\keywords{Out-of-Distribution Detection, AI Reliability, Image Classification}



\maketitle

\section{Introduction}\label{sec1} 

A fundamental component of application safety is modeling the expected operational domain, which provides boundaries for when it is sensible to trust the application and when it does not. However, it is challenging to define such an operational domain for machine learning programs, especially for visual classifiers based on Deep Neural Networks (DNNs), which could lead to potentially catastrophic consequences in real-world applications. 

Out-of-distribution (OOD) detection methods \citep{hendrycks17baseline,liu2020energy,yang2021generalized,yang2022openood} are developed to determine the application scope of deep models in real-world scenarios, working on the same eventual goal of detecting risk cases that deep models can't give reliable predictions on. There have been plenty of works on designing a criterion to construct the application scope of deep models, including deriving the criterion from the features extracted by the model \citep{Mahalanobis,sun2022knn}, the outputs of the model \citep{hendrycks17baseline, liu2020energy}, the combination of features and outputs \citep{wang2022vim}, and the gradient of the model \citep{huang2021importance}.

Existing OOD detection methods have achieved satisfactory performance on conventional evaluations \citep{huang2021mos,sun2021react,huang2021importance,sun2022dice}, which roughly consider the whole validation dataset of the model as the in-distribution dataset and the dataset that has disjoint labels with respect to the in-distribution dataset as the out-of-distribution dataset. 
We notice that there exists a discrepancy between the conventional evaluation and the essential goal of OOD detection that enhances the reliability and safety of deep models.

\begin{figure*}[htbp]
  \includegraphics[width=\textwidth]{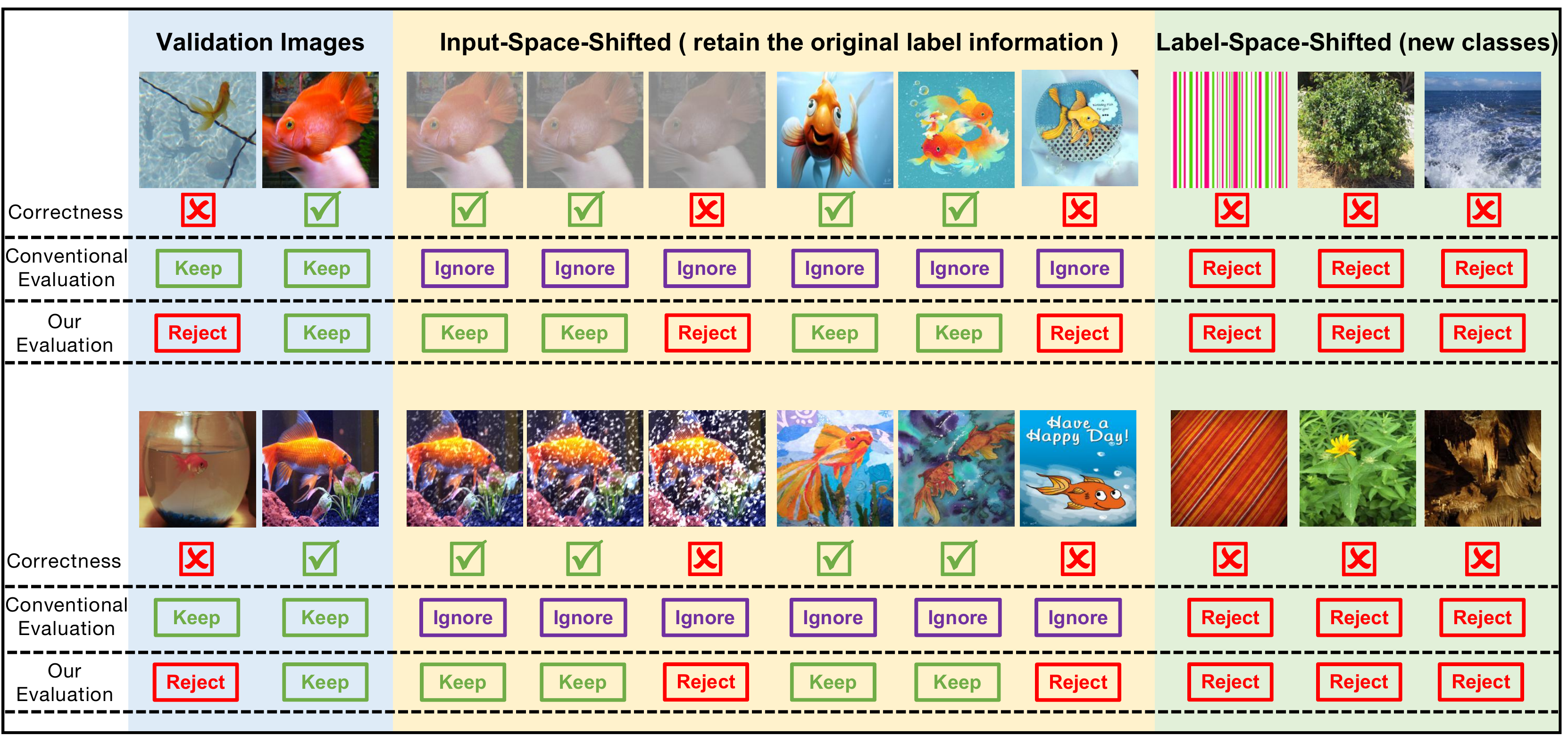}
  \caption{Images in the blue area are validation images of the ResNet-50 pretrained on ImageNet \citep{deng2009imagenet}. Images in the yellow area are input-space-shifted images that retain the label information of the original dataset (including corruptions and stylized versions). Images in the green area are label-space-shifted images that belong to new categories of the model. ``Correctness" means whether the model's prediction matches the human expectation (the ground truth label). Conventional evaluations roughly regard all validation images as in-distribution images and exclusively consider the risk caused by label-space shifts while ignoring the risk caused by input-space shifts. Our human-centric evaluation proposes to reject the risk from both the input space shift and the label space shift and keep making predictions on the images that can be correctly classified by the model.}
  \label{fig:intro}
\end{figure*}

First, the ``distribution" in the ``in-distribution" in the conventional evaluation is defined over the whole validation dataset or the training dataset \citep{hendrycks17baseline, sun2021react} of the deep model, and the conventional evaluations reward OOD detection methods for not detecting the misclassified images in the validation dataset. 
This does not meet humans' safety needs because the misclassified input can also threaten the security of the model and should be rejected rather than kept.
Second, the conventional evaluations exclusively focus on the risk caused by the label-space-shifted inputs \citep{huang2021mos,yang2021generalized} that belong to new categories. However, the input-space shifts \citep{huang2021importance,Imagenet_C,Imagenet_R} where inputs can be corruption-shifted or domain-shifted but retain the label information of the original dataset may also cause risks for classification. 
It is challenging for conventional evaluation to determine ``To what extent input space data variation should be regarded as out of distribution and rejected". Roughly rejecting or adopting all the input-space-shifted images is unreasonable because deep models can reliably handle part of the input-space-shifted images, and the risk comes from the misclassified part.
Fig. \ref{fig:intro} shows that the conventional evaluation may not be suitable to determine which cases humans wish to reject or keep.

In this paper, we appeal to rethink OOD detection from a human-centric perspective, which can provide proper solutions for the above two dilemmas of conventional evaluation.
First, according to human security needs, examples that the OOD detector regards as ``in-distribution" should be reliable examples of the deep model. Thus, the ``distribution" in ``in-distribution" should be defined over the images on which the deep model's prediction meets the human expectation in the training dataset rather than over the whole training dataset or validation dataset like conventional evaluations.
Second, we extend the evaluation of OOD detection to consider both label-space and input-space shifts. Specifically, we propose the OOD detection methods to reject the input on which the model’s prediction mismatches human expectations and keep the input on which the model’s prediction meets the human expectation.

In practice, we use the ground-truth labels annotated by humans of the dataset as alternatives to human expectations.
As shown in Fig. \ref{fig:intro}, from the human-centric perspective, the OOD detector should keep making predictions on images that can be correctly classified by the model even though the input space of the image is shifted but should reject images that models can not make correct predictions on. 
The human-centric evaluation method is more in line with the essential purpose of OOD detection and the safety requirements of humans.

We have conducted extensive empirical evaluations of OOD detection methods with various model architectures, including MLP, CNN, and the recent transformer-based model ViT. Our results lead to two major conclusions. First, the baseline method maximum softmax probability (MSP \citep{hendrycks17baseline}) achieves comparable and even better performance on different distribution-shifted cases when compared to recent popular OOD detection alternatives. 
This finding suggests that the efforts over the years may be overestimated and not have intrinsically addressed human safety concerns, and demonstrates that the need for rethinking in this field may outweigh the need to propose new detection methods.
Second, model architecture and training regime selection matter in OOD detection. Some OOD detection methods almost fail on a certain model but may achieve the best performance on another model. 
This finding reflects ignoring the selection of models and claiming the universality across different models, like the existing OOD detection methods, may not be rigorous.

The main contributions of our paper are summarized as follows:
 \begin{itemize}

  \item[$\bullet$] We provide a novel OOD detection evaluation method from a human-centric perspective that takes whether the model's prediction meets human expectations into account.
 
 \item[$\bullet$] We found that current progress in OOD detection might have been overestimated. Most advanced methods tailor-made for OOD detection are still on par with or even worse than the baseline method maximum softmax probability (MSP).

 \item[$\bullet$] We found that the training regime and model architecture selection matter in OOD detection, which should be regarded as an integral part of the proposed method rather than claiming the proposed method is universal across different models.
 
 \end{itemize}

The rest of the paper is organized as follows. Section II summarizes the literature related to OOD detection methods and evaluation methods. In Section III, we first provide the formulation of the OOD detection task.
Then we introduce the conventional evaluation method and analyze its defects.
After that, we appeal for attention to the discrepancy between the existing evaluation method and the essential goal of the OOD detection task. Moreover, we propose to take the model's performance into account and rethink the OOD detection task from a human-centric perspective.
In Section IV, we first conduct OOD detection experiments with existing methods on different models, which shows that the simple baseline method maximum softmax probability (MSP) beats the other recent methods from the literature. Then, we investigate the influence of model capacity and training strategy on detection performance. Section V gives some conclusive results.

\section{Related work}\label{sec2}

It is well known that a deep neural network can yield unreliable predictions on anomaly inputs from an unknown distribution. When adopted in safety-critical systems such as medical diagnosis or autonomous driving, it is essential to detect examples that the classifier likely fails to make proper predictions. The task of OOD detection is to distinguish the out-of-distribution input (on which the prediction is unreliable) from the in-distribution (ID) input (on which the prediction is reliable). OOD detection has received wide attention because it is critical to ensuring the safety of deep neural networks.

\subsection{OOD Detection Methods} 
Directly estimating the density of the examples can be a natural approach to quantify the uncertainty of input, which explicitly models the distribution of ID examples with probabilistic models and distinguishes the OOD vs. ID examples through the likelihood \citep{nalisnick2018deep,kobyzev2020normalizing,zisselman2020deep,serra2019input,ren2019likelihood, xiao2020likelihood,kirichenko2020normalizing}. However, these methods require expensive costs to train the probabilistic model and perform lag behind the classification-based approaches \citep{huang2021mos}. 

The classification-based approaches use the classifier or a part of the classifier (e.g., feature extractor) to construct a criterion for OOD detection.

Confidence enhancement methods propose retraining the classifier to enhance the sensitivity to OOD examples. Taking advantage of the adversarial training optimization technique, \cite{hein2019relu} endow low confidence to the examples far away from the training data and high confidence to the training data. Moreover, some methods introduce a set of collected OOD examples into the training process to enhance the uncertainty estimation and enforce low confidence around the OOD examples \citep{hendrycks2018deep,papadopoulos2021outlier,chen2021atom}. The correlations between the collected and real OOD examples have a great influence on the performance of these outliers exposure methods \citep{wu2021ngc}. 

Post-hoc methods focus on improving the OOD detection performance with the pre-trained classifiers rather than retraining a model, which is beneficial for adopting OOD detection in real-world scenarios and large-scale settings. These methods derive OOD criterion from different spaces of the deep classifier, e.g., output space \citep{hendrycks17baseline,liu2020energy,ODIN,sun2021react,zhu2022boosting}, feature space \citep{Mahalanobis,sun2022knn}, and gradient space \citep{huang2021importance}. The methods based on the output space start from a simple baseline MSP \citep{hendrycks17baseline}, which hypothesizes that the classifier output a higher maximum softmax probability on the ID example than the OOD example. ODIN \citep{ODIN} introduces a large sufficiently temperature factor and adversarial perturbation to amplify the difference between the softmax probability between the ID and OOD examples. 
\cite{liu2020energy} analyze the limitations of softmax probability in OOD detection and propose to use the  negative energy as a criterion. The negative energy is termed as the energy score, and the examples with low energy scores are regarded as OOD examples, and vice versa. \cite{sun2021react} and \cite{zhu2022boosting} propose to rectify the features of the classifier to improve the detection performance. Virtual-logit Matching (ViM) \citep{wang2022vim} proposes a softmax score which is jointly determined by the feature and the existing logits.
The methods based on the feature space suppose the features of OOD examples should be relatively far away from that of in-distribution classes. \cite{Mahalanobis} model the distribution of feature representations with a mixture of Gaussians and propose using the feature-level Mahalanobis distance as an OOD criterion.
KNN-OOD \citep{sun2022knn} computes the k-th nearest neighbor distance between the feature of the test input and the features of the models' training dataset.
GradNorm \citep{huang2021importance} investigates the gradient space of the classifier and shows that the gradients of the categorical cross-entropy loss can be used as an uncertainty measurement.

\subsection{OOD Evaluation Datasets} 
OOD examples are generally one of two types: i) \textbf{label-space-shifted} examples \citep{van2018inaturalist,xiao2010sun,cimpoi2014Texture,zhou2017places,hendrycks2021nae} and ii) \textbf{input-space-shifted} examples \citep{Imagenet_C,Imagenet_R,SIN}. The label-space-shifted examples belong to a new category that is different from the training dataset, therefore these examples would not be correctly predicted by deep models. The label-space-shifted examples are conventionally used in OOD evaluation. iNaturalist \citep{van2018inaturalist} is a natural fine-grained dataset that contains images whose labels are disjoint from ImageNet-1k. The Scene UNderstanding (SUN) \citep{xiao2010sun} dataset and Places \citep{zhou2017places} dataset are scene recognition datasets that can be used as label-space-shifted datasets against ImageNet. The Describable Textures Dataset \citep{cimpoi2014Texture} is a texture dataset, which can be divided into 47 categories according to human perception. ImageNet-O \citep{hendrycks2021nae} contains anomalies of unforeseen classes which should result in low-confidence predictions and enables us to evaluate the out-of-distribution detection method when the label distribution shifts.
The shifts in input space \citep{Imagenet_C,Imagenet_R,hendrycks2021nae}, where images can be corruption-shifted and domain-shifted while remaining in the same label space, are commonly used to evaluate model robustness and domain generalization performance. ImageNet-C \citep{Imagenet_C} applies different corruptions to the ImageNet validation dataset, including noise, blur, compression, etc. ImageNet-R \citep{Imagenet_R} contains various renditions of ImageNet classes, including art, cartoons, graffiti, sketches, etc.
The conventional OOD detection evaluation generally focuses on the risk caused by the label-space shifts while ignoring the risk from the input-space shifts.

\section{OOD Evaluation}

\subsection{Task Formulation}

Given a deep classifier to solve the classification problem with $K$ classes whose labels are denoted as ${\mathcal Y} = \{1,2,\ldots, K\}$.
Let ${\mathcal X}$ be the input space. Let $f: {\mathcal X} \mapsto {\mathbb R}^K$ represent the pre-trained classifier.
Suppose that the deep classifier can give reliable predictions on the in-distribution data and the distribution of the in-distribution data ${\mathcal D}_{in}$ is denoted as $P_0$. 

OOD detection methods aim to improve the reliability of safety-critical deep models by filtering out the samples for which the model cannot make proper predictions.
Thus, OOD detection tasks commonly determine a reject region ${\mathcal R}$ based on a criterion and a threshold. For any test input $\mathbf{x} \in {\mathcal X}$, the classifier rejects to give a prediction on the test input if $\mathbf{x} \in {\mathcal R}.$
One can use the classifier $f$ or a part of $f$ (e.g., feature extractor) to construct a criterion $T(\mathbf{x}; f)$, where $\mathbf{x}$ is the test input. Then the reject region can be written as $
{\mathcal R} = \{\mathbf{x}: T(\mathbf{x};f) \leq \gamma\}
$, where $\gamma$ is the threshold.

The OOD detection criterion is required to give higher scores for the in-distribution examples that classifiers can make reliable predictions and give lower scores for the out-of-distribution examples that classifiers can not handle.
When considering the out-of-distribution detection in the classification task to ensure the safety of deep neural networks, the output of the classifier can be formulated as follows:
\begin{equation}
    (f,T)(\textbf{x};\gamma):=\begin{cases}
    \textbf{Rejection}, & \text{if} \quad T(\mathbf{x};f) \leq \gamma \\
    f(x), & \text{otherwise}.
    \end{cases}
    \label{Eq:fomulation}
\end{equation}
The alarm is triggered when $T(\mathbf{x};f)$ falls below the threshold $\gamma$. 

\subsection{Conventional OOD Evaluation}\label{Sec:Conventional}

Recently proposed OOD detection methods are commonly assessed on conventional OOD evaluation, which uses the FPR95 and AUROC to measure the detection performance. The whole validation dataset of the deep model is typically considered the in-distribution dataset. The label space of the out-of-distribution dataset is disjointed from the in-distribution dataset.

\textbf{FPR95:} the false positive rate of OOD (negative) examples when the true positive rate of in-distribution (positive) examples is as high as 95$\%$. 
Given a threshold $\gamma$, the confusion matrix for OOD detection can be expressed as follows:
\begin{equation}
    TP(\gamma) =  \sum_{i=1}^{n} (1-y_{ood}) \cdot \mathbb{I}(T(\mathbf{x}_i;f) \geq \gamma ),
\label{Eq:TP}
\end{equation}
\begin{equation}
    FN(\gamma) =  \sum_{i=1}^{n} (1-y_{ood}) \cdot \mathbb{I}(T(\mathbf{x}_i;f) < \gamma ),
\label{Eq:FN}
\end{equation}
\begin{equation}
    FP(\gamma) =  \sum_{i=1}^{n} (y_{ood}) \cdot \mathbb{I}(T(\mathbf{x}_i;f) \geq \gamma ),
\label{Eq:FP}
\end{equation}
\begin{equation}
    TN(\gamma) =  \sum_{i=1}^{n} (y_{ood}) \cdot \mathbb{I}(T(\mathbf{x}_i;f) < \gamma ),
\label{Eq:TN}
\end{equation}

where $y_{ood}$ is the OOD label which is 1 for the OOD example and 0 for the ID example. In the conventional OOD evaluation, the OOD example is of the unseen category that the model hasn’t been trained on. $\mathbb{I}(Event)$ represent the indicator function:
\begin{equation}
    \mathbb{I}(Event):=\begin{cases}
    1, & \text{if the Event is True} \\
    0, & \text{otherwise}.
    \end{cases}
    \label{Eq:indicator}
\end{equation}

The \textbf{t}rue \textbf{p}ositive \textbf{r}ate (TPR) can be computed as:
\begin{equation}
    TPR(\gamma) = \frac{TP(\gamma)}{TP(\gamma)+FN(\gamma)},
\label{Eq:TPR}
\end{equation}
The \textbf{f}alse \textbf{p}ositive \textbf{r}ate (FPR) can be computed as:
\begin{equation}
    FPR(\gamma) = \frac{FP(\gamma)}{FP(\gamma)+TN(\gamma)}.
\label{Eq:FPR}
\end{equation}

\textbf{AUROC:} the area under the receiver operating characteristic curve (ROC), which is the plot of TPR vs. FPR.

\subsection{Defects of Conventional OOD Evaluation}

Existing evaluations roughly regard all the validation images which share the same label space with the training dataset as the in-distribution dataset and regard images of new classes as out-of-distribution images. 
However, the essential goal of the OOD task is to improve the model's safety by detecting examples in which models can not give reliable predictions.
Thus, there exists a discrepancy between the OOD task and the conventional OOD evaluation.

\begin{figure}[htbp]
\centering
\includegraphics[width=0.46\textwidth]{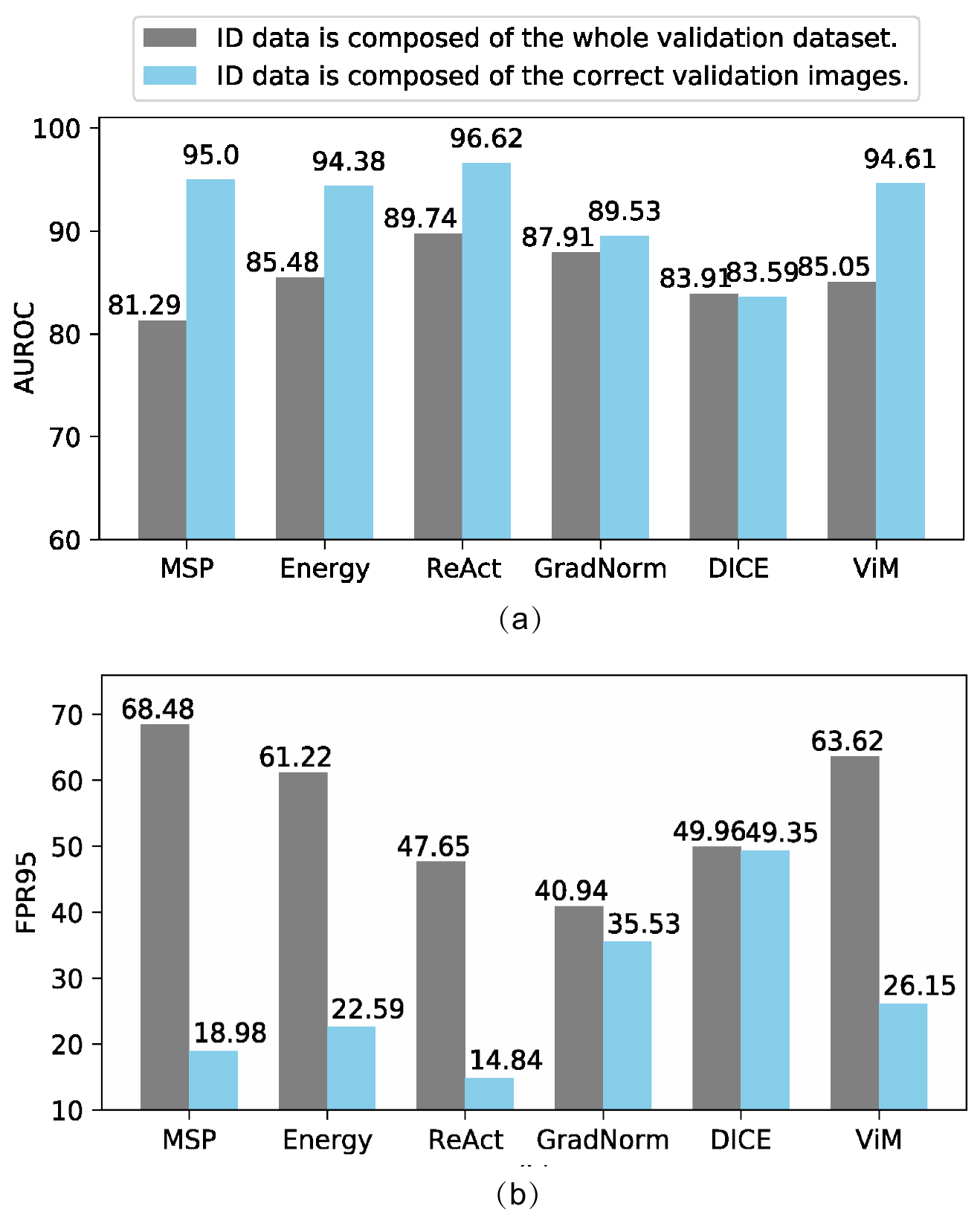}
\caption{(a) The AUROC for different methods (Higher is better) on ResNet-50. (b) The FPR95 for different methods (lower is better) on ResNet-50. The Resnet-50 is pre-trained on the ImageNet dataset. The performance is averaged on four commonly used OOD datasets (iNaturalist, Places, SUN, and Textures). The gray bar indicates that we regard all Imagenet validation examples as ID examples. The blue bar indicates that we regard only the correctly classified examples in the Imagenet validation as ID examples.}
\label{img:Val_ID}
\end{figure}

First, regarding all the validation images (including misclassified and correctly classified examples) as the in-distribution dataset is contrary to the human security needs that in-distribution images should be reliably classified. 
According to the goal of OOD detection, the detection method should assign lower scores to the unrecognizable examples and reject to make predictions on these examples.
However, existing evaluation penalizes the OOD detection method for assigning lower scores for misclassified examples in the validation dataset and rewards the OOD detection method for assigning higher scores for these misclassified examples.
The discrepancy between the commonly stated purpose of OOD detection and the evaluation method may lead to the underestimation of some OOD detection methods. 
As shown in Fig. \ref{img:Val_ID}, the range of in-distribution data has a significant impact on OOD detection performance.
When considering all the validation images of ImageNet \citep{deng2009imagenet} as in-distribution images, the baseline method MSP \citep{hendrycks17baseline} lags behind other methods. However, when eliminating the influence of the misclassified images, MSP shows comparable performance with other recent methods. Specifically, the performance of the MSP in FPR95 is 27.54$\%$ worse than that of GradNorm when regarding all validation images as in-distribution images but surpasses GradNorm by 16.55$\%$ when only regarding the correctly classified images as in-distribution images.

Second, the conventional OOD evaluation method roughly considers the risk caused by label-space-shifted examples whose label does not belong to the categories of the training dataset and ignores the risk caused by input-space-shifted cases. 
It is unreasonable to regard input-space-shifted examples as in-distribution examples because these examples may lead to failure of the classification and cause potential risks.
It is also not advisable to arbitrarily treat all the input-space-shifted examples as OOD examples. Consider the following scenario: the input-space-shifted data is constructed by adding small Gaussian noises. Deep models can provide reliable prediction on this input-space-shifted data, and the OOD detector should not reject this data.

\subsection{Human-centric OOD Evaluation}

Aware of defects of conventional OOD detection evaluation, we propose to rethink the OOD detection methods from a human-centric perspective.
That is, the ``distribution" in ``in-distribution" and ``out-of-distribution" should indicate the distribution of the images on which the model's prediction meets the humans' expectations in the training dataset rather than the distribution of the whole training dataset. We denote this distribution as $P_0$.  
``In-distribution" means the test example is drawn from the distribution $P_0$ where the deep models can provide reliable predictions on.
Test examples that deviate away from the distribution $P_0$ should be rejected.
From a human-centric perspective, the examples on which the model's prediction mismatches the human's expectation should be rejected, and examples on which the model's prediction meets the human's expectation should be kept.

The ground-truth labels of images in the test datasets are commonly annotated by humans.
In practice, we use the ground truth label of the image as an alternative to human expectation. 
The confusion matrix can be reformulated as follows with a given threshold $\gamma$:
\begin{equation}
    TP_{cor}(\gamma) =  \sum_{i=1}^{n} (y_{cor}) \cdot \mathbb{I}(T(\mathbf{x}_i;f) \geq \gamma ),
\label{Eq:TP2}
\end{equation}
\begin{equation}
    FN_{cor}(\gamma) =  \sum_{i=1}^{n} (y_{cor}) \cdot \mathbb{I}(T(\mathbf{x}_i;f) < \gamma ),
\label{Eq:FN2}
\end{equation}
\begin{equation}
    FP_{cor}(\gamma) =  \sum_{i=1}^{n} (1-y_{cor}) \cdot \mathbb{I}(T(\mathbf{x}_i;f) \geq \gamma ),
\label{Eq:FP2}
\end{equation}
\begin{equation}
    TN_{cor}(\gamma) =  \sum_{i=1}^{n} (1-y_{cor}) \cdot \mathbb{I}(T(\mathbf{x}_i;f) < \gamma ),
\label{Eq:TN2}
\end{equation}

where $y_{cor}$ is the label of whether the image can be correctly classified. $y_{cor}$ is 1 for the example that the prediction of the model is consistent with the ground-truth label.

We propose to measure the error rate at different thresholds. The threshold is estimated on the correctly classified images in the training dataset (in-distribution images). The error comes from two aspects, one is the assignment of recognizable examples to the reject region ($FN_{cor}$), and the other is the failure to reject unrecognizable examples ($FP_{cor}$).

The \textbf{d}etection \textbf{e}rror \textbf{r}ate (DER) can be fomulated as:
\begin{footnotesize}
\begin{equation}
    DER(\gamma) = \frac{FN_{cor}(\gamma)+FP_{cor}(\gamma)}{TP_{cor}(\gamma)+FN_{cor}(\gamma)+FP_{cor}(\gamma)+TN_{cor}(\gamma)}.
\label{Eq:DER}
\end{equation}
\end{footnotesize}

\begin{algorithm}[t]
    \caption{Human-Centric Evaluation.}
    \begin{algorithmic}
    \State Require: Classifier $f$, in-distribution training dataset $\mathcal{D}_{in}=\{\boldsymbol{x}_i,y_i\}_{i=1}^N$, a test dataset $\mathcal{D}_{test}$, an OOD criterion $T(\boldsymbol{x};f)$ and percentage $p$.
    \State $\blacktriangleright$ \textbf{Estimate the reject region} :
    \State ID$\_${Scores} = [\;]
    \For {$i = 1,2...,N$}
        \State {score = $T(\boldsymbol{x}_i;f)$}
        \State {Add score to ID$\_${Scores}}
        \EndFor
    \State {$\gamma$=quantile(ID$\_${Scores},$\; p$)}    
    \State $\blacktriangleright$ \textbf{Measure the DER on $\mathcal{D}_{test}$} :
    \vspace{0.1cm}
    \State {$DER(\gamma) = \frac{FN_{cor}(\gamma)+FP_{cor}(\gamma)}{TP_{cor}(\gamma)+FN_{cor}(\gamma)+FP_{cor}(\gamma)+TN_{cor}(\gamma)}$}
    \State {Lower DER indicates better OOD detection performance.} 
    \end{algorithmic}
    \label{algorithm:Human-Centric}
\end{algorithm}

Our evaluation method is outlined in Alg. \ref{algorithm:Human-Centric}. 
It is worth noting that our human-centric OOD evaluation method differs from the conventional evaluation method in two aspects. 
First, the data used to estimate the reject region is different. OOD detection is a single-sample testing task, and the reject region should be determined by known in-distribution examples. We regard correctly classified examples in the training dataset as in-distribution examples and choose the threshold according to the scores of these examples. The conventional evaluation estimate the rejection threshold based on the model's validation dataset in which the misclassified images may lead to the underestimation of the reject region.

\begin{figure}[htbp]
\centering
\includegraphics[width=0.475\textwidth]{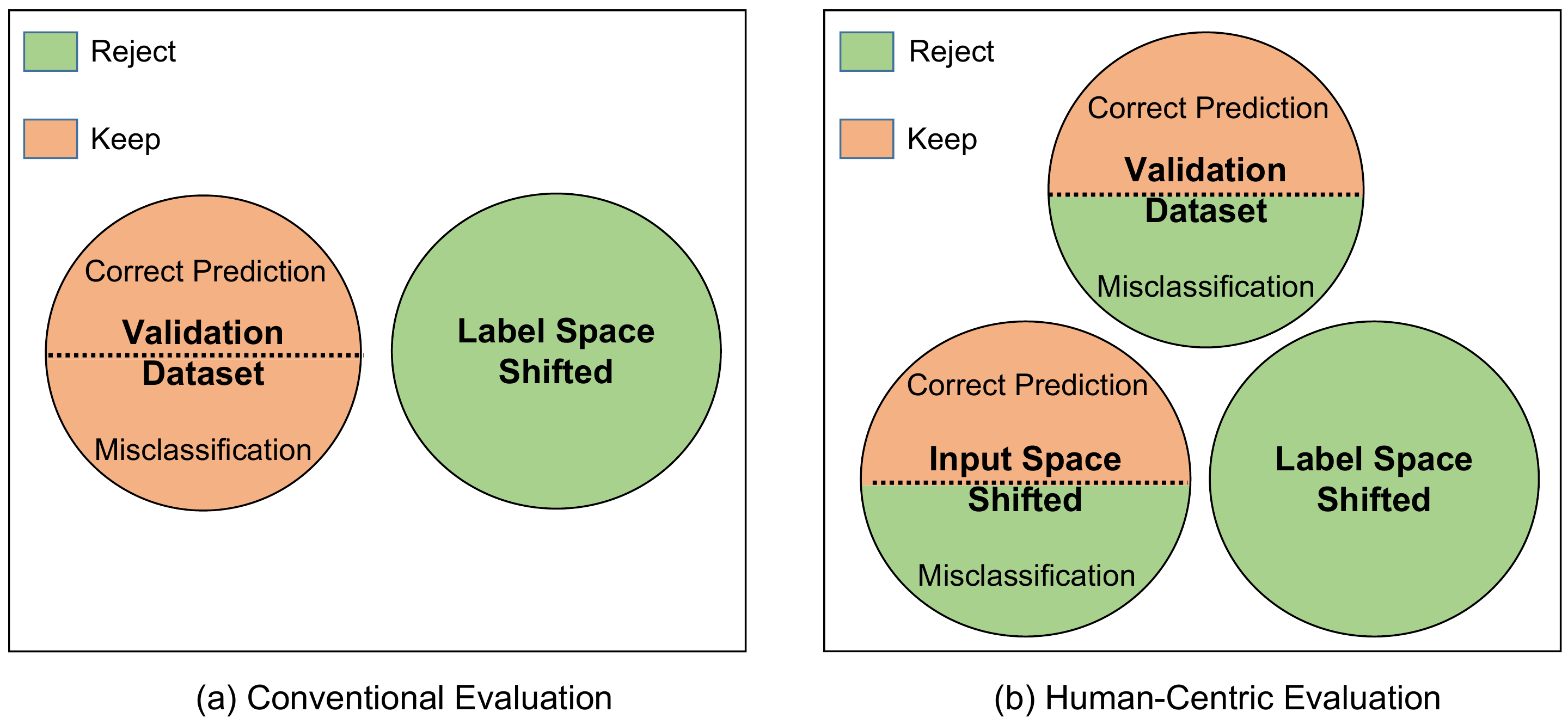}
\caption{(a) The conventional evaluation aims to evaluate the performance of the OOD detection method on distinguishing the label-space-shifted images from the validation dataset (which is regarded as an in-distribution dataset). (b) Our human-centric evaluation takes the model's performance into account and aims to evaluate the performance of the OOD detection method in rejecting the misclassified images and keeping the correctly predicted images.}
\label{img:Scope}
\end{figure}

Second, the criterion for judging whether the detection result is correct is different. The conventional evaluation requires the OOD detection methods to assign lower scores for the images with unseen categories of the model, while our human-centric evaluation requires OOD detection methods to assign lower scores for the images that the model can not provide a reliable prediction. Fig. \ref{img:Scope} illustrates the difference between our human-centric evaluation and the conventional evaluation. 
Our human-centric OOD evaluation broad the scope of evaluating the OOD detection performance from exclusively detecting risks caused by label-space-shifted images to detecting risks caused by multi-type distribution-shifted cases. 

\section{Experiments}\label{exp}

\subsection{Implementation}
We evaluate different OOD detection methods with human-centric evaluation in Alg. \ref{algorithm:Human-Centric}.
We name the DER (in Eq. (\ref{Eq:DER})) whose threshold $\gamma$ is chosen at the 95th and 99th percentile of scores of the correctly classified images in the training dataset (in-distribution data) of the deep model as DER95 and DER99. The threshold of DER95 means that 95$\%$ of in-distribution examples' scores are higher than this threshold. An OOD detection method should assign higher scores than the threshold to examples for which the model can make reliable predictions and lower scores than the threshold to examples that the model can not properly handle.
All experiments in this paper are run on Tesla V100. 

\textbf{Algorithms.} We evaluate nine OOD detection methods in our experiments:
 \begin{itemize}

  \item[$\bullet$] \textbf{MSP} score \citep{hendrycks17baseline} uses the maximum softmax probabilities as the criterion score and supposes deep models assign higher probabilities to in-distribution examples.
  
 \item[$\bullet$] \textbf{Mahalanobis} score \citep{Mahalanobis} computes the minimum Mahalanobis distance between the feature of the test example and the class-wise centroids.

  \item[$\bullet$] \textbf{KL-Matching} \citep{hendrycks2019scaling} uses the minimum KL-divergence between the softmax and the mean class-conditional distributions as the OOD indicator.
 
  \item[$\bullet$] \textbf{Energy} score \citep{liu2020energy} proposes an energy function that maps the logit outputs to a scalar through a convenient logsumexp operator. Examples with higher energy are considered OOD examples.

  \item[$\bullet$] \textbf{ReAct} \citep{sun2021react} proposes to rectify the activation before calculating the Energy score.

  \item[$\bullet$] \textbf{GradNorm} \citep{huang2021importance} is an OOD detection method utilizing information extracted from the gradient space.

  \item[$\bullet$] \textbf{KNN} \citep{sun2022knn}  demonstrates the efficacy of the non-parametric nearest-neighbor distance for OOD detection.

  \item[$\bullet$] \textbf{ViM} \citep{wang2022vim} considers both the class-agnostic score from feature space and the In-Distribution (ID) class-dependent logits in OOD detection.

  \item[$\bullet$] \textbf{DICE} \citep{sun2022dice} ranks weights of the model based on a measure of contribution and selectively uses the most salient weights to derive the output for OOD detection.

 \end{itemize}

\begin{table*}[htbp]
\begin{center}
\caption{OOD detection performance ( DER99 $\|$ DER95 ) on different models that were pre-trained on the ImageNet training dataset. A lower detection error rate (DER99 and DER95) indicates better performance. All methods are post hoc and can be directly used for pre-trained models. We consider 8 test datasets, including the validation dataset of ImageNet, 3 input-space-shifted datasets, and 4 label-space-shifted datasets. ``Average" refers to the average performance on 8 datasets. Due to the page width limits, in this table, ``Mahalanobis score" is abbreviated as ``Maha.", ``KL Matching" is abbreviated as ``KLM" and ``GradNorm" is abbreviated as ``Grad.".
The best results are in Bold. All entries are percentages.}
\scalebox{0.672}{
\begin{tabular}{ccccccccccc}
\toprule  
\toprule  
 & Method       & Textures    & iNaturalist & Places      & SUN         & ImageNet-Val  & ImageNet-C  & ImageNetV2  & ImageNet-R  & Average     \\ \toprule  
\toprule  
  
\multirow{15}{*}{\rotatebox{90}{VGG19}} & MSP & 39.50$\|$22.22 & 20.14$\|$9.50 & \textbf{46.52}$\|$\textbf{24.29} & 43.35$\|$22.55 & \textbf{19.99}$\|$\textbf{20.72} & \textbf{25.77}$\|$\textbf{23.34} & \textbf{25.70}$\|$\textbf{23.98} & \textbf{33.70}$\|$28.10 & \textbf{31.83}$\|$21.84 \\ \cmidrule{2-11}
      & Maha.  & 98.53$\|$95.25 & 99.99$\|$99.91 & 99.97$\|$99.87 & 99.98$\|$99.87 & 27.88$\|$29.99 & 50.66$\|$50.96 & 40.48$\|$41.40 & 66.43$\|$66.47 & 72.99$\|$72.97 \\ \cmidrule{2-11}    
      & Energy       & 34.10$\|$16.37 & 20.50$\|$5.87  & 51.88$\|$24.76 & 41.72$\|$\textbf{17.49} & 24.09$\|$23.54 & 32.23$\|$27.91 & 32.59$\|$28.20 & 35.41$\|$\textbf{26.20} & 34.07$\|$21.29 \\ \cmidrule{2-11}    
      & KLM & 41.83$\|$20.16 & 23.07$\|$8.87  & 57.11$\|$30.10 & 55.72$\|$27.81 & 21.77$\|$21.28 & 28.24$\|$24.03 & 28.29$\|$24.53 & 36.08$\|$27.82 & 36.51$\|$23.07 \\ \cmidrule{2-11}    
      & ReAct        & \textbf{32.29}$\|$\textbf{15.05} & \textbf{20.04}$\|$\textbf{5.79}  & 51.74$\|$24.98 & \textbf{41.61}$\|$18.04 & 24.07$\|$23.48 & 32.12$\|$27.71 & 32.58$\|$28.17 & 35.15$\|$26.29 & 33.70$\|$\textbf{21.19} \\ \cmidrule{2-11}    
      & Grad.     & 51.28$\|$28.94 & 50.83$\|$27.96 & 78.29$\|$56.77 & 65.50$\|$42.77 & 26.72$\|$27.04 & 48.37$\|$45.95 & 38.52$\|$36.55 & 60.94$\|$52.69 & 52.56$\|$39.83 \\ \cmidrule{2-11}    
      & KNN          & 98.56$\|$95.35 & 99.97$\|$99.60 & 99.95$\|$99.45 & 99.98$\|$99.76 & 27.78$\|$29.88 & 50.56$\|$49.46 & 40.35$\|$41.22 & 66.12$\|$64.18 & 72.95$\|$72.66 \\ \cmidrule{2-11}    
      & ViM          & 81.31$\|$53.88 & 98.86$\|$90.08 & 98.13$\|$87.62 & 98.99$\|$91.66 & 26.44$\|$26.02 & 46.49$\|$36.52 & 38.63$\|$33.92 & 57.89$\|$41.21 & 68.34$\|$57.61 \\ \cmidrule{2-11}    
      & DICE         & 74.10$\|$44.17 & 85.83$\|$56.15 & 91.63$\|$72.74 & 84.65$\|$59.68 & 27.26$\|$28.28 & 50.35$\|$49.97 & 39.78$\|$39.39 & 65.88$\|$63.43 & 64.93$\|$51.73 \\ \toprule  
\toprule  
        
    \multirow{15}{*}{\rotatebox{90}{RN50}}  & MSP & 41.45$\|$24.59 & 23.70$\|$10.49 & \textbf{45.11}$\|$24.01 & 42.43$\|$22.03 & \textbf{18.78}$\|$\textbf{20.42} & \textbf{25.34}$\|$\textbf{23.50} & \textbf{24.35}$\|$\textbf{23.27} & \textbf{31.58}$\|$28.79 & \textbf{31.59}$\|$22.14 \\ \cmidrule{2-11}    
     & Maha. & 66.86$\|$36.15 & 99.11$\|$93.76 & 99.66$\|$96.79 & 99.69$\|$96.72 & 25.85$\|$27.86 & 51.14$\|$45.77 & 39.08$\|$39.96 & 63.46$\|$60.11 & 68.11$\|$62.14 \\  \cmidrule{2-11}    
     & Energy      & 48.39$\|$24.06 & 35.83$\|$11.33 & 59.32$\|$31.67 & 52.86$\|$25.72 & 22.79$\|$23.18 & 31.74$\|$26.85 & 31.61$\|$27.42 & 34.27$\|$29.45 & 39.60$\|$24.96 \\ \cmidrule{2-11}    
     & KLM & 40.64$\|$20.48 & 23.32$\|$8.12  & 53.52$\|$25.62 & 50.58$\|$22.84 & 19.98$\|$21.00 & 26.52$\|$23.84 & 26.59$\|$24.01 & 34.32$\|$28.56 & 34.43$\|$21.81 \\ \cmidrule{2-11}    
     & ReAct       & 43.24$\|$17.71 & \textbf{20.14}$\|$\textbf{5.50}  & 49.73$\|$\textbf{22.19} & \textbf{41.80}$\|$\textbf{16.71} & 22.93$\|$22.93 & 33.05$\|$26.55 & 31.97$\|$27.20 & 35.22$\|$28.55 & 34.76$\|$\textbf{20.92} \\ \cmidrule{2-11}    
     & Grad.    & 51.95$\|$34.31 & 39.25$\|$20.41 & 72.07$\|$51.04 & 59.09$\|$38.47 & 25.55$\|$26.87 & 48.81$\|$45.26 & 37.92$\|$37.55 & 52.53$\|$44.75 & 48.40$\|$37.33 \\ \cmidrule{2-11}    
     & KNN         & 89.61$\|$71.15 & 99.53$\|$96.19 & 99.16$\|$94.36 & 99.52$\|$95.86 & 25.76$\|$27.27 & 48.11$\|$38.73 & 38.67$\|$38.58 & 62.36$\|$56.28 & 69.22$\|$63.47 \\ \cmidrule{2-11}    
     & ViM         & \textbf{14.29}$\|$\textbf{5.59}  & 71.16$\|$25.73 & 81.62$\|$49.37 & 81.69$\|$45.79 & 24.23$\|$23.50 & 28.02$\|$26.82 & 35.32$\|$29.37 & 36.91$\|$\textbf{27.58} & 46.66$\|$29.22 \\ \cmidrule{2-11}    
     & DICE        & 63.90$\|$45.82 & 59.80$\|$37.18 & 83.52$\|$65.86 & 71.47$\|$52.23 & 25.79$\|$27.55 & 51.41$\|$51.20 & 38.93$\|$39.76 & 59.60$\|$55.57 & 56.80$\|$46.90 \\ \toprule  
\toprule  

\multirow{15}{*}{\rotatebox{90}{NASNetLarge}} & MSP & 46.99$\|$32.16 & 35.62$\|$22.81 & 48.43$\|$31.58 & 45.14$\|$29.51 & \textbf{15.44}$\|$\textbf{19.30} & 22.12$\|$23.87 & 21.53$\|$23.42 & 35.34$\|$33.20 & 33.83$\|$26.98 \\ \cmidrule{2-11}  
     & Maha. & \textbf{22.85}$\|$\textbf{11.86}  & \textbf{10.18}$\|$\textbf{4.11}  & 42.12$\|$22.79 & 38.43$\|$\textbf{18.79} & 17.05$\|$21.82 & 21.87$\|$26.92 & 22.47$\|$25.06 & 25.15$\|$28.52  & \textbf{25.02}$\|$\textbf{19.98} \\ \cmidrule{2-11} 
     & Energy      & 87.41$\|$72.80  & 92.00$\|$78.44 & 94.51$\|$82.66 & 92.80$\|$77.50 & 18.84$\|$21.03 & 32.29$\|$31.01 & 29.47$\|$30.00 & 51.41$\|$47.14  & 62.34$\|$55.07 \\ \cmidrule{2-11} 
     & KLM & 32.77$\|$21.10  & 21.46$\|$13.20 & \textbf{38.21}$\|$\textbf{22.81} & \textbf{35.17}$\|$20.38 & 15.81$\|$19.82 & \textbf{21.12}$\|$\textbf{23.37} & \textbf{20.80}$\|$\textbf{22.88} & 30.70$\|$28.81  & 27.00$\|$21.55 \\ \cmidrule{2-11} 
     & ReAct       & 88.60$\|$64.38  & 89.99$\|$61.33 & 93.29$\|$76.38 & 88.05$\|$66.25 & 18.77$\|$20.89 & 33.63$\|$30.70 & 30.07$\|$29.46 & 53.70$\|$45.72  & 62.01$\|$49.39 \\ \cmidrule{2-11} 
     & Grad.    & 98.69$\|$96.65  & 99.80$\|$99.37 & 99.95$\|$99.40 & 99.55$\|$98.31 & 18.79$\|$20.71 & 34.38$\|$35.32 & 30.06$\|$31.25 & 54.79$\|$54.81  & 67.00$\|$66.98 \\ \cmidrule{2-11} 
     & KNN         & 32.32$\|$16.88  & 16.78$\|$6.80  & 48.12$\|$26.54 & 44.73$\|$22.46 & 17.33$\|$21.93 & 21.95$\|$26.04 & 23.10$\|$25.24 & \textbf{24.10}$\|$\textbf{26.50}  & 26.78$\|$20.77 \\ \cmidrule{2-11} 
     & ViM         & 29.27$\|$14.29  & 14.33$\|$6.15  & 50.06$\|$29.48 & 45.98$\|$24.83 & 17.03$\|$21.02 & 21.92$\|$25.29 & 22.94$\|$24.43 & 24.73$\|$26.52  & 28.28$\|$21.50 \\ \cmidrule{2-11} 
     & DICE        & 99.40$\|$97.62  & 99.76$\|$99.51 & 99.90$\|$99.17 & 99.38$\|$98.30 & 18.66$\|$20.66 & 34.31$\|$35.34 & 30.01$\|$31.10 & 54.80$\|$54.82  & 67.04$\|$67.07 \\ \toprule  
\toprule  
     
\multirow{15}{*}{\rotatebox{90}{Inception-v4}}  & MSP & 41.03$\|$25.98 & 28.25$\|$15.60 & 42.99$\|$\textbf{24.33} & 39.12$\|$\textbf{21.59} & \textbf{16.64}$\|$\textbf{20.35} & 22.29$\|$24.20 & 21.67$\|$23.27 & 31.31$\|$30.16 & 30.41$\|$23.18 \\ \cmidrule{2-11} 
     & Maha. & 34.54$\|$14.45  & 27.84$\|$\textbf{6.97}  & 73.86$\|$41.41 & 71.97$\|$37.29 & 20.03$\|$21.63 & 27.63$\|$25.93 & 28.95$\|$26.70 & 37.34$\|$27.36  & 40.27$\|$25.22 \\ \cmidrule{2-11} 
     & Energy      & 63.81$\|$46.24  & 63.22$\|$41.98 & 72.05$\|$51.59 & 68.62$\|$47.79 & 19.81$\|$21.91 & 30.45$\|$28.53 & 28.12$\|$27.58 & 40.54$\|$$\|$35.79 & 48.33$\|$37.68 \\ \cmidrule{2-11} 
     & KLM & 33.62$\|$21.84  & 20.50$\|$12.38 & \textbf{41.18}$\|$24.84 & \textbf{37.68}$\|$21.79 & 16.89$\|$20.36 & \textbf{22.21}$\|$\textbf{23.68} & \textbf{21.61}$\|$\textbf{22.75} & \textbf{29.70}$\|$28.53  & \textbf{27.92}$\|$\textbf{22.02} \\ \cmidrule{2-11} 
     & ReAct       & 49.41$\|$30.37  & 37.02$\|$20.24 & 63.26$\|$42.31 & 56.42$\|$36.11 & 19.40$\|$21.38 & 28.26$\|$26.71 & 27.02$\|$26.12 & 36.70$\|$31.75  & 39.69$\|$29.37 \\ \cmidrule{2-11} 
     & Grad.    & 96.65$\|$93.14  & 98.73$\|$96.73 & 99.70$\|$98.84 & 99.12$\|$97.19 & 21.37$\|$23.55 & 40.59$\|$41.45 & 32.68$\|$33.94 & 57.84$\|$57.71  & 68.33$\|$67.82 \\ \cmidrule{2-11} 
     & KNN         & 44.08$\|$16.93  & 34.38$\|$8.12  & 67.84$\|$36.67 & 67.18$\|$32.06 & 19.65$\|$21.99 & 26.02$\|$25.79 & 26.94$\|$27.05 & 31.74$\|$25.62  & 40.00$\|$24.75 \\ \cmidrule{2-11} 
     & ViM         & \textbf{26.13}$\|$\textbf{12.70} & \textbf{22.64}$\|$9.39  & 66.25$\|$44.38 & 60.21$\|$37.34 & 19.44$\|$21.21 & 25.88$\|$25.83 & 27.48$\|$26.00 & 29.77$\|$\textbf{25.21}  & 34.73$\|$25.26 \\ \cmidrule{2-11} 
     & DICE        & 99.86$\|$$\|$99.11 & 99.98$\|$99.93 & 99.98$\|$99.80 & 99.94$\|$99.66 & 21.37$\|$23.36 & 40.62$\|$41.50 & 32.68$\|$33.82 & 57.93$\|$57.97  & 69.04$\|$69.39 \\     
    
\toprule  
\toprule  
      
\end{tabular} \label{tab:performance1}}
\end{center}
\end{table*}

\begin{table*}[htbp]
\begin{center}
\caption{Continued for Table 1. OOD detection performance ( DER99 $\|$ DER95 ) on different models that were pre-trained on the ImageNet training dataset. All entries are percentages.}
\scalebox{0.672}{
\begin{tabular}{ccccccccccc}
\toprule  
\toprule  
 & Method       & Textures    & iNaturalist & Places      & SUN         & ImageNet-Val  & ImageNet-C  & ImageNetV2  & ImageNet-R  & Average     \\ \toprule  
\toprule  
  
\multirow{15}{*}{\rotatebox{90}{SENet154}}     & MSP & 35.60$\|$20.64 & 22.52$\|$10.88 & 41.85$\|$23.09 & 38.58$\|$19.63 & \textbf{15.97}$\|$\textbf{19.98} & 21.82$\|$\textbf{24.07} & 21.27$\|$23.36 & 32.11$\|$31.53 & 28.71$\|$21.65 \\ \cmidrule{2-11} 
     & Maha. & 14.27$\|$7.07  & \textbf{10.31}$\|$\textbf{3.44}  & 42.56$\|$19.60 & 38.64$\|$14.84 & 17.97$\|$21.32 & 23.30$\|$27.20 & 24.36$\|$24.53 & 25.06$\|$27.88 & 24.56$\|$\textbf{18.23} \\ \cmidrule{2-11} 
     & Energy      & 45.32$\|$30.71 & 36.56$\|$19.19 & 56.51$\|$37.59 & 53.81$\|$33.30 & 18.14$\|$21.68 & 24.99$\|$26.39 & 24.93$\|$26.52 & 33.36$\|$32.54 & 36.70$\|$28.49 \\ \cmidrule{2-11} 
     & KLM & 28.07$\|$15.34 & 15.94$\|$7.61  & \textbf{36.35}$\|$19.12 & 34.07$\|$15.93 & 16.12$\|$20.26 & \textbf{21.32}$\|$24.09 & \textbf{20.88}$\|$\textbf{22.93} & 29.60$\|$29.16 & 25.29$\|$19.30 \\ \cmidrule{2-11} 
     & ReAct       & 21.79$\|$11.38 & 12.19$\|$5.73  & 40.84$\|$23.47 & \textbf{34.04}$\|$16.97 & 17.81$\|$21.23 & 22.83$\|$26.02 & 23.40$\|$24.85 & 28.07$\|$29.11 & 25.12$\|$19.84 \\ \cmidrule{2-11} 
     & Grad.    & 97.38$\|$95.69 & 99.92$\|$99.74 & 99.92$\|$99.62 & 99.85$\|$99.52 & 20.12$\|$22.15 & 37.41$\|$38.40 & 31.76$\|$33.06 & 58.10$\|$58.23 & 68.06$\|$68.30 \\ \cmidrule{2-11} 
     & KNN         & 20.64$\|$9.63  & 12.25$\|$4.23  & 40.31$\|$\textbf{18.05} & 37.74$\|$\textbf{14.04} & 17.87$\|$21.44 & 22.85$\|$26.53 & 23.42$\|$24.67 & 25.02$\|$\textbf{27.81} & 24.79$\|$18.27 \\ \cmidrule{2-11} 
     & ViM         & \textbf{12.57}$\|$\textbf{6.52}  & 10.52$\|$3.67  & 41.47$\|$20.32 & 36.15$\|$14.47 & 17.87$\|$21.30 & 23.19$\|$27.47 & 23.84$\|$24.75 & \textbf{24.86}$\|$27.82 & \textbf{23.81}$\|$18.29 \\ \cmidrule{2-11} 
     & DICE        & 99.80$\|$99.38 & 99.88$\|$99.78 & 99.99$\|$99.85 & 99.99$\|$99.86 & 20.10$\|$22.14 & 37.45$\|$38.29 & 31.77$\|$33.12 & 58.09$\|$58.21 & 68.40$\|$68.86 \\
\toprule  
\toprule  
\multirow{15}{*}{\rotatebox{90}{MobileNetV2}} & MSP & 59.43$\|$32.93 & \textbf{43.30}$\|$17.21 & \textbf{65.49}$\|$\textbf{35.86} & 64.09$\|$34.62 & \textbf{23.90}$\|$\textbf{20.47} & \textbf{33.42}$\|$\textbf{24.20} & \textbf{31.30}$\|$\textbf{24.50} & \textbf{40.41}$\|$29.10 & \textbf{45.17}$\|$\textbf{27.36} \\ \cmidrule{2-11}
     & Maha. & 84.65$\|$64.33 & 99.95$\|$99.48 & 99.91$\|$99.13 & 99.93$\|$99.38 & 30.53$\|$32.20 & 57.38$\|$56.73 & 43.92$\|$44.78 & 69.54$\|$67.88 & 73.23$\|$70.49 \\ \cmidrule{2-11}
     & Energy      & 55.94$\|$29.52 & 47.03$\|$16.31 & 67.15$\|$38.88 & 61.29$\|$32.20 & 27.43$\|$25.42 & 37.46$\|$28.66 & 37.07$\|$30.67 & 43.74$\|$28.29 & 47.14$\|$28.74 \\ \cmidrule{2-11}
     & KLM & 63.28$\|$30.16 & 48.93$\|$16.81 & 76.92$\|$44.11 & 78.21$\|$43.22 & 26.40$\|$22.07 & 37.41$\|$25.42 & 35.35$\|$26.48 & 48.53$\|$30.16 & 51.88$\|$29.80 \\ \cmidrule{2-11}
     & ReAct       & 53.10$\|$25.57 & 44.86$\|$\textbf{14.51} & 66.19$\|$37.43 & 59.61$\|$30.40 & 27.40$\|$25.35 & 37.30$\|$28.48 & 36.99$\|$30.59 & 43.55$\|$\textbf{27.77} & 46.12$\|$27.51 \\ \cmidrule{2-11}
     & Grad.    & 45.25$\|$23.95 & 45.53$\|$21.96 & 68.04$\|$42.77 & \textbf{53.70}$\|$\textbf{29.25} & 29.43$\|$28.88 & 50.80$\|$41.74 & 41.66$\|$38.00 & 57.63$\|$43.20 & 49.01$\|$33.69 \\ \cmidrule{2-11}
     & KNN         & 92.39$\|$81.95 & 99.91$\|$99.51 & 99.90$\|$99.09 & 99.96$\|$99.60 & 30.59$\|$32.11 & 57.32$\|$55.41 & 43.94$\|$44.64 & 69.06$\|$66.40 & 73.68$\|$71.41 \\ \cmidrule{2-11}
     & ViM         & \textbf{33.63}$\|$\textbf{11.49} & 96.66$\|$78.54 & 96.97$\|$83.05 & 97.10$\|$84.13 & 29.48$\|$27.56 & 43.33$\|$30.55 & 42.00$\|$35.70 & 55.17$\|$31.19 & 61.79$\|$47.78 \\ \cmidrule{2-11}
     & DICE        & 52.48$\|$31.12 & 56.75$\|$34.21 & 76.37$\|$53.41 & 60.98$\|$37.98 & 30.22$\|$30.86 & 56.01$\|$53.14 & 43.33$\|$42.22 & 64.67$\|$56.26 & 55.10$\|$42.40 \\       
\toprule  
\toprule  
\multirow{15}{*}{\rotatebox{90}{Swin-B}}  & MSP         & 34.63$\|$20.05 & 10.51$\|$4.30  & \textbf{38.76}$\|$20.41 & 35.69$\|$17.54 & \textbf{13.66}$\|$19.38 & 23.26$\|$42.20 & \textbf{18.93}$\|$\textbf{23.67} & 31.33$\|$37.43 & 25.85$\|$23.12 \\  \cmidrule{2-11} 
     & Maha. & 35.41$\|$15.21 & 1.81$\|$0.45   & 47.17$\|$20.31 & 44.83$\|$16.02 & 15.76$\|$21.28 & 27.40$\|$39.22 & 21.86$\|$25.60 & 31.35$\|$40.38 & 28.20$\|$22.31 \\ \cmidrule{2-11} 
     & Energy      & 46.38$\|$30.60 & 12.69$\|$7.03  & 51.70$\|$36.10 & 44.99$\|$29.89 & 15.71$\|$20.24 & 30.17$\|$43.71 & 23.71$\|$27.26 & 33.45$\|$35.71 & 32.35$\|$28.82 \\ \cmidrule{2-11} 
     & KLM & 31.21$\|$17.04 & 7.99$\|$2.66   & 39.40$\|$\textbf{18.01} & \textbf{35.09}$\|$15.72 & 14.45$\|$19.93 & 26.63$\|$42.28 & 19.66$\|$23.71 & 31.82$\|$38.11 & 25.78$\|$22.18 \\ \cmidrule{2-11} 
     & ReAct       & 38.42$\|$23.63 & 6.75$\|$3.20   & 44.40$\|$28.76 & 37.94$\|$23.03 & 15.64$\|$20.55 & 29.95$\|$44.97 & 22.84$\|$26.95 & 31.83$\|$35.70 & 28.47$\|$25.85 \\ \cmidrule{2-11} 
     & Grad.    & 96.15$\|$94.17 & 89.20$\|$86.92 & 94.07$\|$92.74 & 91.86$\|$90.15 & 15.43$\|$\textbf{18.01} & 27.88$\|$\textbf{32.01} & 25.33$\|$27.81 & 39.16$\|$39.20 & 59.88$\|$60.14 \\ \cmidrule{2-11} 
     & KNN         & 46.52$\|$19.98 & 10.70$\|$1.18  & 53.58$\|$24.80 & 53.82$\|$21.12 & 15.74$\|$21.24 & \textbf{21.66}$\|$36.13 & 22.01$\|$25.12 & \textbf{29.49}$\|$37.94 & 29.94$\|$22.87 \\ \cmidrule{2-11} 
     & ViM         & \textbf{25.48}$\|$\textbf{11.45} & \textbf{0.96}$\|$\textbf{0.24}   & 41.16$\|$18.79 & 36.38$\|$\textbf{12.98} & 15.86$\|$21.24 & 31.69$\|$40.84 & 21.99$\|$25.85 & 32.21$\|$41.42 & \textbf{25.71}$\|$\textbf{21.60 }\\ \cmidrule{2-11} 
     & DICE        & 44.24$\|$23.30 & 45.71$\|$24.59 & 52.01$\|$30.44 & 44.88$\|$25.60 & 15.55$\|$19.15 & 33.36$\|$41.95 & 23.85$\|$25.63 & 35.60$\|$\textbf{35.58} & 36.90$\|$28.28 \\ \toprule  
\toprule  
\multirow{15}{*}{\rotatebox{90}{ViT-B16}} & MSP         & 35.41$\|$19.91 & 8.16$\|$3.23   & 39.12$\|$20.02 & 36.12$\|$18.40 & \textbf{12.82}$\|$\textbf{18.17} & \textbf{17.20}$\|$\textbf{23.94} & \textbf{18.33}$\|$\textbf{22.88} & 31.63$\|$38.82 & 24.85$\|$20.67 \\ \cmidrule{2-11} 
     & Maha. & \textbf{19.79}$\|$\textbf{9.59 } & \textbf{1.18}$\|$\textbf{0.41}   & 43.36$\|$18.91 & 32.21$\|$\textbf{11.43} & 15.13$\|$20.93 & 21.12$\|$30.98 & 21.80$\|$25.80 & 32.42$\|$41.43 & 23.38$\|$19.94 \\ \cmidrule{2-11} 
     & Energy      & 21.91$\|$11.65 & 2.66$\|$0.91   & \textbf{33.49}$\|$\textbf{17.35} & \textbf{25.58}$\|$12.54 & 15.28$\|$20.58 & 21.86$\|$29.77 & 21.42$\|$25.25 & 35.98$\|$44.46 & \textbf{22.27}$\|$20.31 \\ \cmidrule{2-11} 
     & KLM & 28.67$\|$16.97 & 5.03$\|$2.08   & 37.45$\|$19.13 & 33.74$\|$16.68 & 13.99$\|$18.84 & 18.93$\|$24.37 & 19.76$\|$23.64 & 34.03$\|$39.69 & 23.95$\|$20.17 \\ \cmidrule{2-11} 
     & ReAct       & 22.11$\|$11.90 & 2.35$\|$0.82   & 36.56$\|$18.43 & 28.27$\|$12.74 & 15.29$\|$20.29 & 21.14$\|$29.02 & 21.17$\|$25.09 & 33.74$\|$42.84 & 22.58$\|$\textbf{20.14} \\ \cmidrule{2-11} 
     & Grad.    & 24.82$\|$17.07 & 3.22$\|$1.80   & 36.37$\|$23.31 & 28.47$\|$17.68 & 14.73$\|$18.19 & 21.14$\|$27.12 & 21.31$\|$24.25 & 34.33$\|$39.87 & 23.05$\|$21.16 \\ \cmidrule{2-11} 
     & KNN         & 24.54$\|$12.34 & 2.05$\|$0.68   & 42.93$\|$20.02 & 37.41$\|$14.70 & 15.71$\|$21.75 & 22.31$\|$31.81 & 22.10$\|$27.70 & 34.30$\|$43.23 & 24.27$\|$20.73 \\ \cmidrule{2-11} 
     & ViM         & 21.77$\|$10.60 & 1.87$\|$0.58   & 46.04$\|$22.18 & 33.66$\|$14.28 & 14.86$\|$19.11 & 21.00$\|$28.65 & 21.75$\|$24.31 & \textbf{29.85}$\|$\textbf{38.81} & 23.85$\|$19.81 \\ \cmidrule{2-11} 
     & DICE        & 92.82$\|$80.83 & 99.75$\|$98.53 & 99.60$\|$97.30 & 99.12$\|$96.00 & 14.61$\|$17.33 & 24.53$\|$32.70 & 24.19$\|$27.24 & 39.14$\|$39.30 & 61.72$\|$61.15 \\ 
    
\toprule  
\toprule  
      
\end{tabular} \label{tab:performance2}}
\end{center}
\end{table*}

\textbf{Datasets.} We evaluate the OOD detection methods by detecting the risk caused by both the input-space-shift and label-space-shift cases. We regard the correctly classified images in the training dataset of ImageNet \citep{deng2009imagenet} as the in-distribution dataset. We consider \textbf{8} test datasets, including Textures \citep{cimpoi2014Texture} which contains 47 categories of textures, iNaturalist \citep{van2018inaturalist} which contains 5089 natural fine-grained categories, Places \citep{zhou2017places} which contains more than 400 different types of scene environments, SUN \citep{xiao2010sun} which contains 397 categories of scene environments, ImageNetV2 \citep{recht2019imagenetv2} which is independent of existing models, ImageNet-C \citep{Imagenet_C} which contains different corruption versions (weather, noise, brightness, et al.) of ImageNet, ImageNet-R \citep{Imagenet_R} which contains different stylized versions (art, sketch, cartoon, et al.) of ImageNet and ImageNet-Val which represents the validation dataset of ImageNet.
The label $y_{cor}$ (in Eq. (\ref{Eq:TP2})) for the datasets Textures, iNaturalist, Places, and SUN is $0$ because the label space of these datasets is disjoint from the training dataset of the deep model and the deep model can not make correct predictions. The label $y_{cor}$ for the datasets ImageNet-C, ImageNet-R, ImageNetV2, and ImageNet-Val indicates whether the model's prediction meets the ground-truth label.

\textbf{Model Architectures.} 
We extensively conduct experiments on \textbf{45} deep models with various model architectures, from Multilayer Perceptron (MLP) based model ResMLP \citep{ResMLP}, classical CNN models (VGG \citep{simonyan2015vgg}, ResNet \citep{he2016deep}, DenseNet \citep{huang2018densenet}, MobileNet-V2 \citep{sandler2019mobilenetv2}, NASNet \citep{zoph2018learning}, the InceptionV4 \citep{szegedy2016inceptionv4} and the squeeze-and-excitation network \citep{hu2019senet}) to the latest transformer models (ViT \citep{dosovitskiy2020imageVIT}, Swin \citep{liu2021Swin} and DeiT \citep{DeiT}). The pre-trained CNN models are provided by torchvision 0.10.0 \citep{pytorch}, and other pre-trained models are provided by timm 0.4.12 \citep{rw2019timm}.

\begin{figure*}[htbp]
\centering
\includegraphics[width=0.99\textwidth]{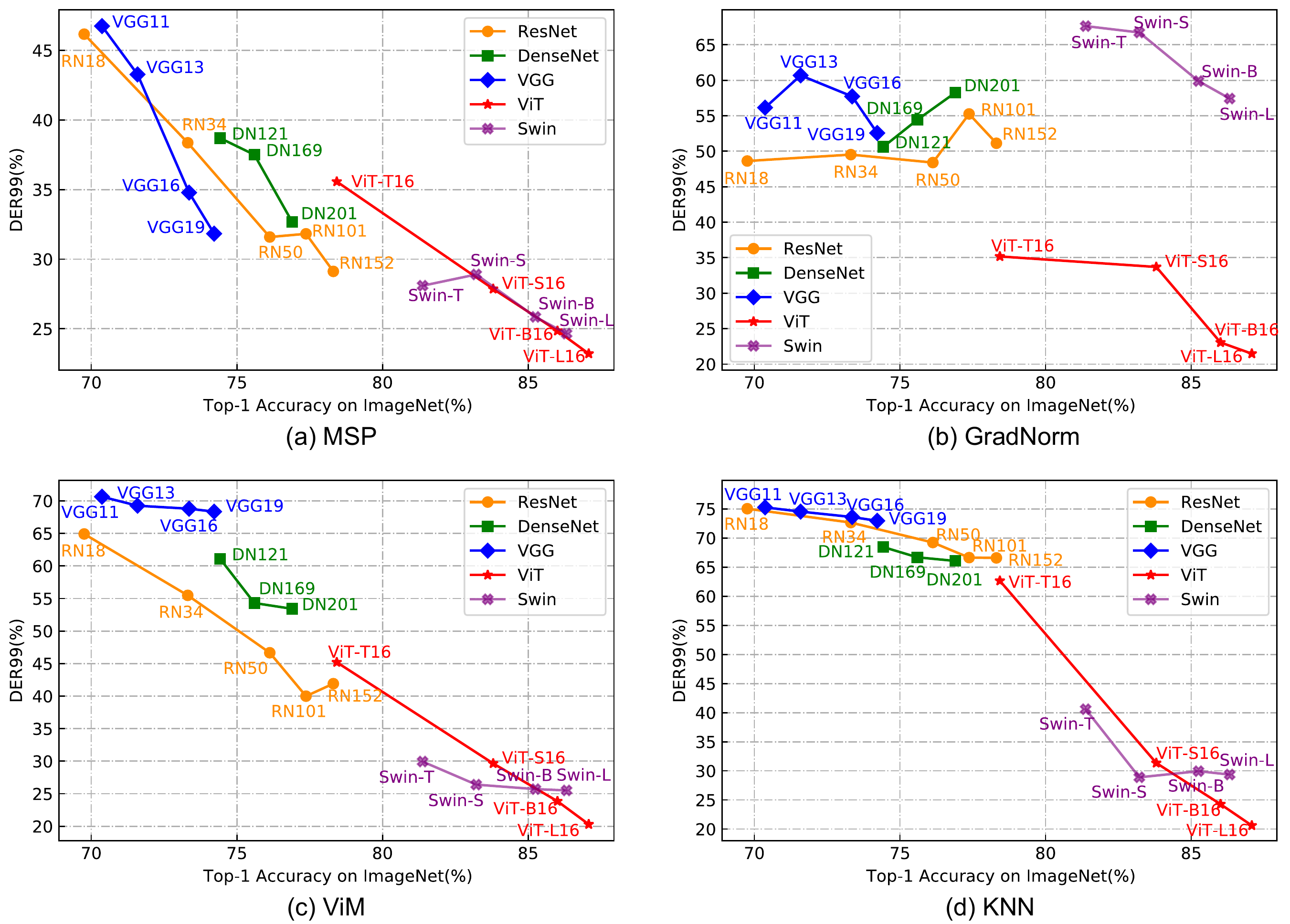}
\caption{The influence of the model's capacity on OOD detection methods (Lower DER99 indicates better detection performance). We choose VGG, ResNet, DenseNet, ViT, and Swin with different capacities. The performance is averaged on eight test datasets. The horizontal axis represents the accuracy of different models.}
\label{img:capacity}
\end{figure*}

\subsection{Results on Different Architectures}
Prevalent OOD detection evaluations exclusively focus on the risk caused by a part of distribution shifts (label space shift) and ignore the risk caused by the input space shift. For the first time, we evaluate the existing detection methods on a variety of models, and different distribution-shift cases, hoping to draw general methodological conclusions.

Tab. \ref{tab:performance1} and Tab. \ref{tab:performance2} show the detection performance results on different model architectures. For each model architecture, we report the OOD detection performance of different methods on each test dataset and the average performance on eight datasets.
We draw two main conclusions from our results:

\textbf{From a human-centric perspective, no method can consistently outperform MSP by a significant margin.} Research on OOD detection originates from a simple baseline MSP  \citep{hendrycks17baseline, yang2021generalized} and receives increased attention. However, our experiment results reveal that the development of OOD detection in the past five years may be overestimated. In Tab. \ref{tab:performance1}, the baseline method MSP achieves state-of-the-art performance in the average DER99 on VGG19, ResNet-50, and MobileNet-V2.
When considering VGG19 as the classifier, MSP surpasses the recently proposed ReAct and ViM by 1.87$\%$ and 36.5$\%$ in DER99, respectively.
On other models, although MSP is not the best one, it only lags behind the best method to a small extent. Specifically, when considering Swin-B as the classifier, MSP lags behind the best method ViM by 0.14$\%$ but surpasses ReAct and GradNorm by 2.62$\%$ and 34.03$\%$, respectively.

\textbf{Model architecture matters in detection methods.}
Besides ignoring the performance of the model, the existing methods also ignore the influence of the model architecture. The existing methods are always evaluated on several models and then claim to be universal across different models.
We conduct experiments on a variety of model architectures and find that the impact of model architecture on detection methods can be fatal. For instance, KNN surpasses MSP by 4.9$\%$ with SENet154 in the average DER99 but its performance drops dramatically with ResNet50. That is to say, although KNN performs well with the SENet154, it's unreliable to use KNN as the OOD detection algorithm to improve the security of the ResNet50 model. Similarly, ViM performs poorly with VGG19, and the detection error rate (DER99) is even over 98$\%$ when evaluated on the iNaturalist dataset, but it can achieve the best performance in the Swin model. Contrary to prior works that arbitrarily claim their method is universal and performs well across different model architectures, we suggest the proposed OOD detection method takes the model selection strategy into consideration.

We do not claim that any of these recently proposed methods cannot possibly improve the detection performance, but getting detection performance improvements is challenging and should regard the model architecture as an integral part of the proposed method. The newly proposed OOD detection methods should remain true to their original aspiration, aiming to improve the safety and reliability of the deep models rather than distinguishing the difference between datasets. The future research direction of OOD detection can be to design detection methods that are more in line with human needs and guarantee the safety of the deep models facing different distribution-shifted cases.

\subsection{The Influence of Model Capacity}

Generally, the model capacity is positively correlated with the classification accuracy. For example, the classification accuracy of models ResNet-18, ResNet-34, ResNet-50, ResNet-101, and ResNet-152 increases. Then, how does the model's capacity affect the performance of OOD detection methods? In this subsection, we investigate the influence of model capacity on the OOD detection methods.

Fig. \ref{img:capacity} illustrates the performance of OOD detection methods in models with different capacities.
We found an interesting phenomenon that there is no monotonous relationship between the performance of detection methods and model capacity. When using the DenseNet model as the classifier, the performance of GradNorm becomes worse with the increase in model capacity. 
When using ViM to conduct OOD detection, its performance with the model ResNet101 is 6.67$\%$ better than with the smaller model ResNet50 and 1.92$\%$ better than with the larger model ResNet152.
Additionally, KNN in the Swin-S(mall), Swin-B(ase), Swin-L(arge) performs better than in the Swin-T(iny), but KNN in the Swin-B(ase) performs worse than in the Swin-S(mall). 
On the whole, the accuracy of the model and the OOD detection performance are positively correlated. Models with higher accuracy tend to achieve a lower OOD detection error rate (DER99). The model ViT-L16 which obtains the best Top-1 accuracy also performs the best in OOD detection. The model ResNet-18, whose Top-1 accuracy is the worst, performs poorly in OOD detection. However, there are also some anomalies. ViT-T16 can achieve better Top-1 accuracy than ResNet-101. MSP and ViM perform better in ResNet-101 than in ViT-T16, while GradNorm and KNN perform worse in ResNet-101 than in ViT-T16.

\subsection{The Influence of Training Regimes}

\subsubsection{Do Adversarial Robustness Contributes to OOD Detection?}

\begin{figure}[htbp]
\centering
\includegraphics[width=0.495\textwidth]{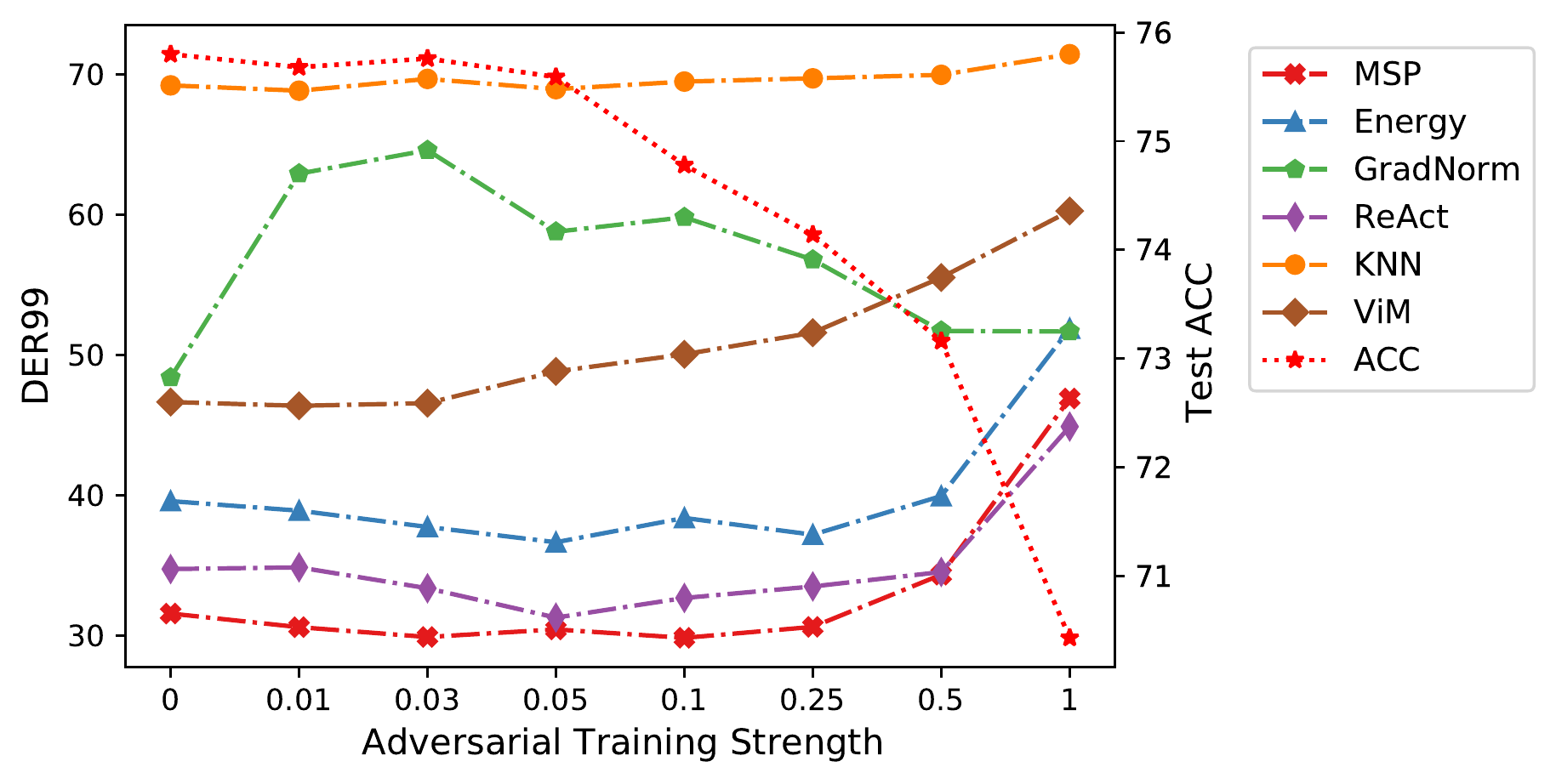}
\caption{The DER99 for different methods (lower is better) on different robust models. The horizontal axis represents the perturbation strength in training the model, e.g. "0.1" represents the robust model trained with $\ell_{2}$ perturbation $\epsilon=0.1$. The red dotted line indicates the test accuracy of the robust model. The performance is averaged on eight test datasets.}
\label{img:Adv}
\end{figure}

Adversarial training~\citep{salman2020adversarially} is efficient in improving the adversarial robustness and the safety of deep neural networks.
To find out whether adversarial training can improve the OOD detection performance of the model, we conduct experiments in terms of different perturbation strengths  (constraints of $l_2$ norm bound). In Fig. \ref{img:Adv}, we evaluate the OOD detection performance on eight adversarially pre-trained ResNet-50. When the strength of adversarial training is small, it slightly impacts OOD detection. However, when the strength of adversarial training is high, the OOD detection performance of this model decrease dramatically. Specifically, the DER99 of MSP on the model adversarially trained with strength $\epsilon=1$ is 15.32$\%$ worse than that on the normal model. MSP almost performs the best on different robust models.

\subsubsection{Do Training Strategy Contributes to OOD Detection?}

\begin{figure}[htbp]
\centering
\includegraphics[width=0.495\textwidth]{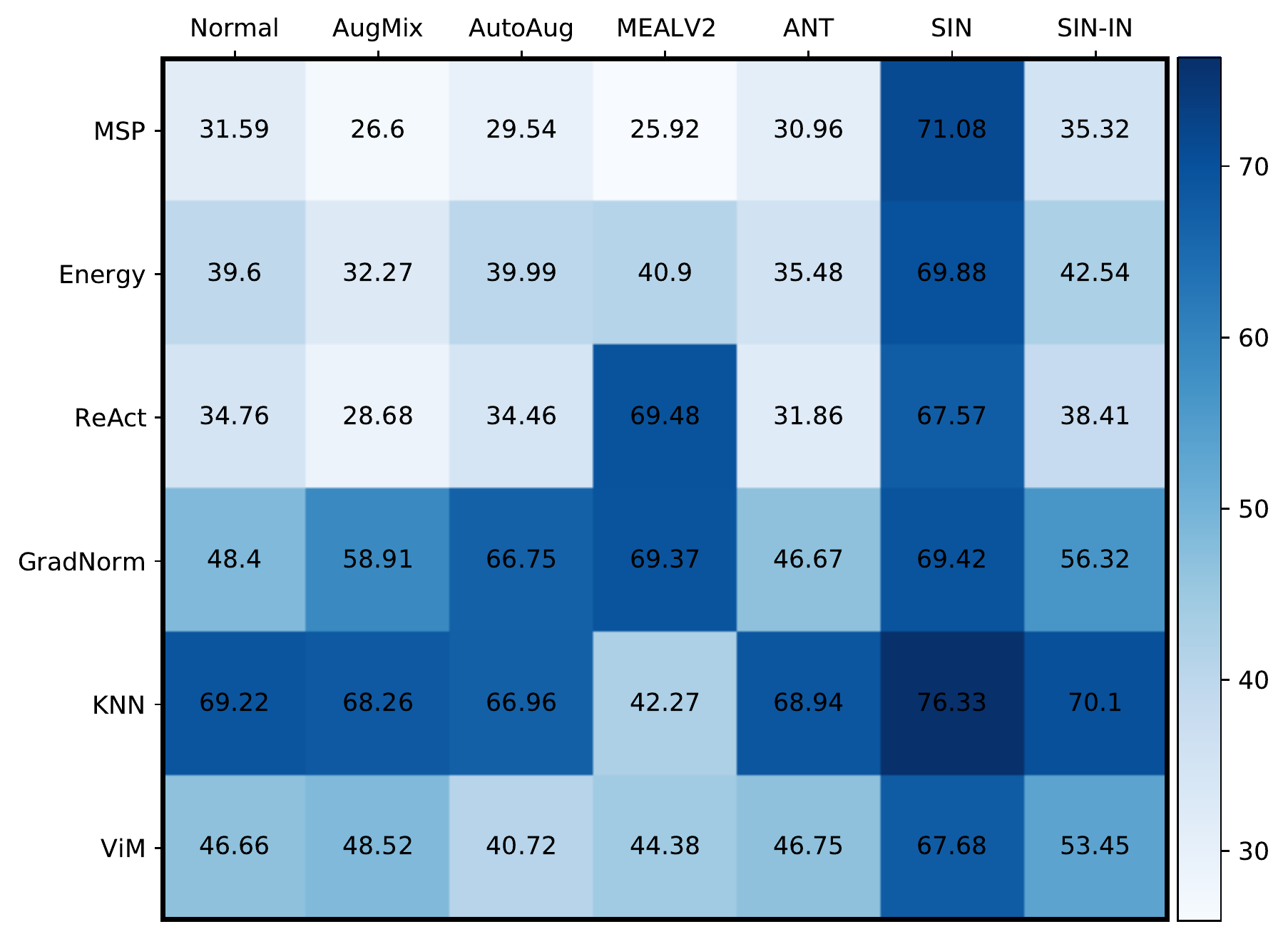}
\caption{The influence of the training strategies on OOD detection. Lighter color indicates better performance (lower DER99). The performance is averaged on eight test datasets.}
\label{img:training_robust}
\end{figure}

Various studies have focused on training strategies to increase the model's performance in classification tasks. To find out the influence of the training strategies on OOD detection, we further consider different strategies, including training with Styled ImageNet (SIN)~\citep{SIN}, training with the mixture of Styled and natural ImageNet (SIN-IN)~\citep{SIN}, training with the augmentation method Augmix~\citep{hendrycks*2020augmix}, training with the augmentation method AutoAugment \citep{lim2019fastAutoAugment}, training with additive Gaussian and Speckle noise (ANT~\citep{rusak2020simpleANT}), and training with the knowledge distillation method MEALV2 \citep{shen2020mealv2} that achieves 80$\%$+ Top-1 accuracy on ResNet-50. As shown in Fig. \ref{img:training_robust}, the most beneficial training strategies for different OOD detection methods are different. MSP and KNN perform the best in the model trained with the MEALV2 strategy. GradNorm outperforms ViM in the ANT strategy-trained model, whereas ViM outperforms GradNorm in the AutoAug strategy-trained model.

\subsubsection{Do Knowledge Distillation Contributes to OOD Detection?}

\begin{figure}[htbp]
\centering
\includegraphics[width=0.495\textwidth]{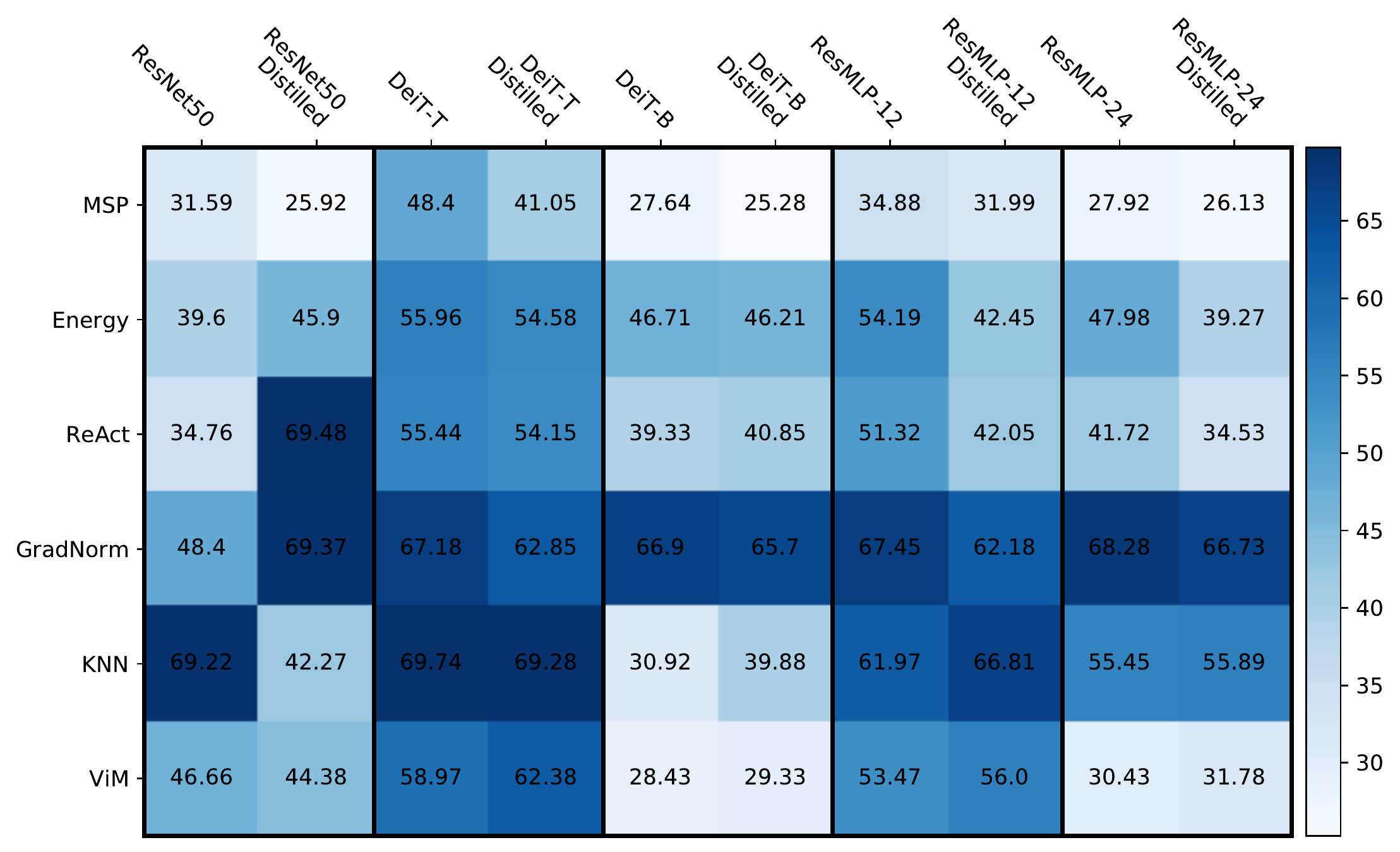}
\caption{The influence of the distillation on OOD detection. Lighter color indicates better performance (lower DER99). The performance is averaged on eight test datasets.}
\label{img:distill}
\end{figure}

In Fig. \ref{img:distill}, we investigate whether knowledge distillation can improve the detection performance of different methods. We consider the CNN-based model ResNet-50, MLP-based models ResMLP-12 and ResMLP-24, transformer-based models DeiT-T and DeiT-B, and their distilled versions. We find that MSP's performance on the distilled model is better than that on the normal model. The influence of the distillation on other methods is inconsistent in different models. For instance, the performance of KNN improves by 26.95$\%$ when using the distilled ResNet-50 instead of the normal ResNet-50, while its performance degrades by 8.96$\%$ when using the distilled DeiT-B instead of the normal DeiT-B. 

These results shed light on the fact that the model architecture and training regime should also be considered when designing the OOD detection algorithm.

\section{Conclusion}\label{sec13}
Current deep learning systems yield uncertain and unreliable predictions when facing distribution-shifted examples. Various OOD detection methods have been proposed to improve the reliability and safety of deep models in real-world applications over the years. 
This work does not propose a novel detection method but analyzes the defects of the conventional evaluation and proposes a human-centric evaluation method that is more in line with the essential goal of OOD detection to investigate how practical existing OOD detection methods are in aiding the safety problem of human needs.
We have conducted extensive experiments on 9 OOD detection methods with 45 different models. Our results draw two major conclusions. First, the progress over the years may not have resulted in much performance gain compared with the baseline method. Second, model architectures and training regimes matter in OOD detection and should be considered integral when designing new detection methods. 
We hope our findings motivate researchers to rethink OOD detection from a human-centric perspective and develop OOD detection methods that can reject risks both from the label space shifts and input space shifts.

\backmatter





\bmhead{Acknowledgments}

This work was supported in part by the Fundamental Research Funds for the Central Universities and by Alibaba Group through Alibaba Research Intern Program.

\section*{Declarations}

\begin{itemize}
\item Funding

This work was supported in part by the Fundamental Research Funds for the Central Universities and by Alibaba Group through Alibaba Research Intern Program.

\item Competing interests

Financial interests: Yao Zhu received a Ph.D. stipend from Zhejiang University. Yuefeng Chen, Xiaodan Li, Rong Zhang, and Hui Xue received salaries from the Alibaba group. Xiang Tian and Yaowu Chen received salaries from Zhejiang University. Bolun Zheng receives a salary from Hangzhou Dianzi University.

Non-financial interests: none.

\item Ethics approval 

Not applicable.

\item Consent to participate

Not applicable.

\item Consent for publication

Not applicable.

\item Availability of data and materials

The datasets analyzed during the current study are available in the ImageNet repository \citep{deng2009imagenet}, \href{https://www.image-net.org/about.php}{https://www.image-net.org/about.php}.

\item Code availability 

We share some codes in the link below to evaluate the performance of different detection methods with our human-centric evaluation. We plan to open-source more codes for the community in the future. We hope our finds can provide insights for future research on OOD detection.

\href{https://drive.google.com/file/d/1avkZrzMoDjqQPoBL6MkYW_javVGrOzln/view?usp=share_link }{https://drive.google.com/file/d/1avkZrzMoDj \\
qQPoBL6MkYW$\_$javVGrOzln/view?usp=shar \\
e$\_$link}

\item Authors' contributions

All authors contributed to the research conception and design. Methodology: [Yao Zhu], [Yuefeng Chen]; Material preparation: [Bolun Zheng], [Rong Zhang]; Formal analysis and investigation: [Xiang Tian], [Rongxin Jiang]; Writing - original draft preparation: [Yao Zhu]; Writing - review and editing: [Yuefeng Chen], [Xiaodan Li]; Funding acquisition: [Hui Xue], [Yaowu Chen].
\end{itemize}

\bibliography{bibliography}


\end{document}